%% file: MyPaper_main.tex
\documentclass[sigconf]{acmart}

\newcommand{\minisection}[1]{\vspace{5pt}\noindent\textbf{#1.}}
\usepackage{enumitem}
\usepackage{appendix}
\usepackage{setspace}

\AtBeginDocument{%
  \providecommand\BibTeX{{%
    \normalfont B\kern-0.5em{\scshape i\kern-0.25em b}\kern-0.8em\TeX}}}

\copyrightyear{2023}
\acmYear{2023}
\setcopyright{acmlicensed}\acmConference[DAI '23]{The Fifth International Conference on Distributed Artificial Intelligence}{November 30-December 3, 2023}{Singapore, Singapore}
\acmBooktitle{The Fifth International Conference on Distributed Artificial Intelligence (DAI '23), November 30-December 3, 2023, Singapore, Singapore}
\acmPrice{15.00}
\acmDOI{10.1145/3627676.3627686}
\acmISBN{979-8-4007-0848-0/23/11}

\begin{document}

\title{Adaptive Control Strategy for Quadruped Robots in Actuator Degradation Scenarios}

\author{Xinyuan Wu}
\authornote{Both authors contributed equally to this research.}
\affiliation{
  \institution{Shanghai Jiao Tong University}
  \city{Shanghai}
  \country{China}}
\email{wuxinyuan@sjtu.edu.cn}

\author{Wentao Dong}
\authornotemark[1]
\affiliation{
  \institution{Shanghai Jiao Tong University}
  \city{Shanghai}
  \country{China}}
\email{a\_Dong@sjtu.edu.cn}

\author{Hang Lai}
\affiliation{%
  \institution{Shanghai Jiao Tong University}
  \city{Shanghai}
  \country{China}}
\email{laihang99@sjtu.edu.cn}

\author{Yong Yu}
\affiliation{%
 \institution{Shanghai Jiao Tong University}
  \city{Shanghai}
  \country{China}}
\email{yyu@apex.sjtu.edu.cn}

\author{Ying Wen}
\authornote{corresponding author.}
\affiliation{%
 \institution{Shanghai Jiao Tong University}
  \city{Shanghai}
  \country{China}}
\email{Ying.Wen@sjtu.edu.cn}

%%
%% The abstract is a summary of the work to be presented in the
\begin{abstract}
Quadruped robots have strong adaptability to extreme environments but may also experience faults. 
Once these faults occur, robots must be repaired before returning to the task, reducing their practical feasibility. 
One prevalent concern among these faults is actuator degradation, stemming from factors like device aging or unexpected operational events. 
Traditionally, addressing this problem has relied heavily on intricate fault-tolerant design, which demands deep domain expertise from developers and lacks generalizability. 
Learning-based approaches offer effective ways to mitigate these limitations, but a research gap exists in effectively deploying such methods on real-world quadruped robots. 
This paper introduces a pioneering teacher-student framework rooted in reinforcement learning, named \textbf{A}ctuator \textbf{D}egradation \textbf{A}da\textbf{p}tation \textbf{T}ransformer (\textsc{Adapt}), aimed at addressing this research gap. 
This framework produces a unified control strategy, enabling the robot to sustain its locomotion and perform tasks despite sudden joint actuator faults, relying exclusively on its internal sensors. 
Empirical evaluations on the Unitree A1 platform validate the deployability and effectiveness of \textsc{Adapt} on real-world quadruped robots, and affirm the robustness and practicality of our approach. 

\end{abstract}


\begin{teaserfigure}
  \includegraphics[width=\textwidth]{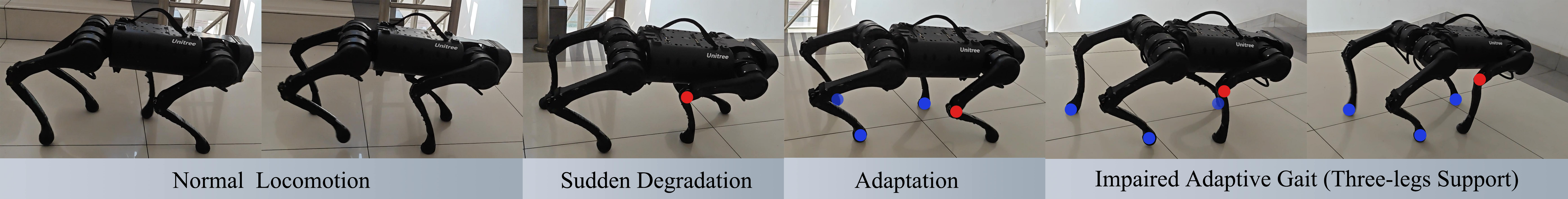}
  % \includesvg[width=\textwidth]{resource/pre.svg}
  \caption{Application of ADAPT facing sudden degradation (indicated by red).}
  \label{fig:pre}
\end{teaserfigure}

\maketitle

\section{Introduction}
\input{_introduction}

\section{RELATED WORK}
\input{_relatedwork}

\section{PRELIMINARIES}
\input{_preliminaries}

\section{Methodology}
\input{_method}

\section{Experiment}
\input{_exp}

\section{Conclusion And Future Work}
\input{_conclusion}

\begin{acks}
We would like to express our sincere gratitude to the National Key R\&D Program of China (2022ZD0114804), Shanghai Municipal Science and Technology Major Project (2021SHZDZX0102) and National Natural Science Foundation of China (62076161) for their support. 
Their contributions have enabled us to complete this work successfully. 
\end{acks}

\bibliographystyle{ACM-Reference-Format}
\bibliography{reference}

\clearpage
\appendix
\input{_appendix}
\end{document}

%% file: _introduction.tex
Legged robots, known for their multifunctionality, have demonstrated remarkable adaptability and flexibility in traversing complex terrains and navigating diverse and unfamiliar environments \cite{miki2022learning, bellicoso2018advances}.
As hardware technology and control algorithms advance, quadruped robots have evolved to possess improved load-bearing capacities, heightened stability, and structural advantages \cite{biswal2021development}.
These advancements have resulted in their growing deployment across diverse domains such as exploration \cite{miki2022learning}, search and rescue \cite{https://doi.org/10.1002/rob.21683}, military applications \cite{kemper2022k9}, and industrial scenarios \cite{bellicoso2018advances}.

% \minisection{Robustness Challenges in Real World}
However, in real-world applications, the robustness of quadruped robots is critically compromised by their susceptibility to various fault scenarios.
While these robots excel in navigating complex and unpredictable environments \cite{biswal2021development}, their adaptability renders them vulnerable to various fault scenarios, such as physical damage, joint malfunctions, and motor aging. 
Such vulnerabilities can have far-reaching implications, leading to actuator degradation and consequent reductions in operational efficiency \cite{liu2022saving, khalastchi2018sensor}. 
Moreover, these faults not only inconvenience operators but, in extreme cases, may also jeopardize the longevity of the robot and pose significant safety risks to humans. 
Therefore, how to deal with such a fault has drawn significant attention from researchers. 

Dealing with the fault in real-time is exceptionally challenging due to obstacles like distance-related accessibility issues and hazardous conditions. 
Additionally, immediate fault rectification often necessitates specialized expertise and tools, which may not be readily available.
As a result, the design of fault-tolerant control algorithms becomes crucial, empowering robots to adapt to varying degrees of malfunction autonomously \cite{cui2022fault, chen2022fault}. 
Implementing such algorithms not only prolongs the robot's operational lifespan but also substantially enhances its reliability. This, in turn, broadens the range of real-world applications where quadruped robots can be effectively deployed.

While traditional fault-tolerant algorithms like Model Predictive Control (MPC) and whole-body control frameworks have been widely adopted, they suffer from concrete drawbacks such as heavy reliance on domain expertise, labor-intensive manual tuning, and limited adaptability to unforeseen environmental and mechanical conditions \cite{di2018dynamic, khalastchi2018sensor, cui2022fault, liu2022saving}. 
These conventional approaches focus on the active detection and confirmation of faults during the robot's operation. If a fault is identified, a pre-engineered alternative control scheme is invoked to mitigate the issue. 
However, the effectiveness of these strategies is intrinsically tied to the designer's comprehensive understanding of the robot's mechanical intricacies, often necessitating meticulous manual adjustments \cite{guo2019robots}. 
Moreover, the rigidity of relying on predefined alternative control schemes becomes a significant constraint in real-world applications. 
Specifically, these methods struggle to adapt to novel environmental or mechanical scenarios, frequently requiring the formulation of separate, customized control strategies for different variants of the same fault \cite{liu2022saving}. 

While learning-based approaches, particularly DRL, offer substantial advantages such as autonomous skill acquisition and reduced dependency on domain expertise, they also present specific challenges, including sample efficiency, algorithmic stability, and the complexity of bridging the simulation-to-reality gap \cite{margolis2022rapid, arulkumaran2017deep, lai2023sim}.
% In contrast, learning-based approaches present a promising alternative.
% \ying{
% Alternative, similarly, conclude the advantages and  concrete challenging and drawbacks of learning-based methods at the first sentence.

% "While learning-based approaches, particularly Deep Reinforcement Learning (DRL), offer substantial advantages such as autonomous skill acquisition and reduced dependency on domain expertise, they also present specific challenges, including sample efficiency, algorithmic stability, and the complexity of bridging the simulation-to-reality gap \cite{margolis2022rapid, arulkumaran2017deep, lai2023sim}." }
This paradigm empowers robots to acquire emergency-response skills independently \cite{guo2019robots}, reducing the dependence on domain knowledge when designing algorithms to some extent \cite{liu2022saving, allard2023online}. 
%Among them, d 
Deep Reinforcement Learning (DRL) has gained prominence in quadruped robot control due to its ability to learn nonlinear control strategies \cite{margolis2022rapid}.
Unlike traditional control methods, DRL does not require precise models and effectively manages high-dimensional state spaces \cite{arulkumaran2017deep}.
To balance exploration and exploitation, reinforcement learning algorithms consistently develop strategies with enhanced adaptability and generalization capabilities \cite{lai2023sim, margolis2022rapid, wu2023learning, yu2023multi}.
Furthermore, the availability of physically accurate modeling simulation environments like IsaacGym \cite{makoviychuk2021isaac} and massively parallel solutions \cite{pmlr-v164-rudin22a} has significantly elevated the efficiency of control strategy learning while easing the transfer from simulation to real-world environments.

While substantial progress has been made in enhancing robot motion performance and enabling cross-terrain capabilities under stable operational conditions \cite{rma, terrain}, there remains a notable gap in exploring actuator degradation fault within the context of DRL for real-world quadruped robots. 
% This research strives to bridge this gap by designing a control algorithm that empowers quadruped robots to adapt to varying degrees of actuator fault in real-world scenarios. 

% In response to this challenge, 
% \ying{after reading above paragraphs still unclear what challenges we are trying to address, because fault robustness is a big question. absolutely, we use the learning-based method, but which challenges our method can address? sample efficiency? algorithmic stability? simulation-to-reality gap? It would be better to narrow down the discussion in previous paragraph to the problem scope or research questions that will be solved in this paper.
% Based on my understating, we are trying to solve the actuator fault problem using teacher-student + decision transformer, which has better sample efficiency or generalization ability?
% } 

In response to bridge the gap, 
we present a pioneering teacher-student training framework that combines transformer-based architectures with reinforcement learning principles, 
% \weinan{add the full name of ADAPT here.}
named \textbf{A}ctuator \textbf{D}egeneration \textbf{A}da\textbf{p}tation \textbf{T}ransformer (\textsc{Adapt}). 
This framework equips the quadruped robot to adapt to joint actuator degradation faults within predetermined thresholds, relying solely on its perceptual abilities. Our proposed framework demonstrates remarkable robustness within simulated environments and showcases the potential for zero-shot transfer to real-world robots.
% \weinan{add more descriptions about the advantages of ADAPT here before the bullets of contributions.}
This learning-based strategy reduces the dependence on deep expertise in the field of robotics or specific robot models while also holding significant potential for acquiring a broader range of skills.

This paper contributes three key aspects to quadruped robot control amidst joint actuator degradation faults.
\begin{itemize}[leftmargin=10pt]
    \item Firstly, we tackle an issue of varying degrees of actuator degradation in an intricate simulation environment. We approach this as a type of cross-embodiment task using reinforcement learning techniques that can handle various actuator degradation scenarios instead of certain pre-defined cases. 
    % \ying{we have never mentioned cross-embodiment in previous text, I guess it may be used to simulate varying degrees of actuator faults. If it is right, make the relationship clear.} 
    \item Secondly, we introduce \textsc{Adapt}, a novel adaptation framework that demonstrates significant generalization potential in experiments. %meticulously designed to enhance the adaptability of quadruped robots facing joint actuator degradation faults. 
    \item Lastly, we successfully deployed \textsc{Adapt} on real-world quadruped robots and impressively demonstrated the capability for zero-shot transferring from simulation to real-world robots. 
\end{itemize}
These contributions collectively pave the way for more resilient and versatile quadruped robot locomotion strategies. 
To the best of our knowledge, this is the first work to successfully deploy a learning-based approach in addressing actuator degradation on real-world quadruped robots. 

%% file: _relatedwork.tex
%是否提及传统控制策略控制四足机器人移动的成果？
% \subsection{DRL for Quadrupedal Locomotion}
\subsection{DRL for Robot Control}
Deep Reinforcement Learning (DRL) holds the potential to reduce the dependence on expertise in robot control by training a policy in simulation and then transferring it to real world. However, it's challenging to directly apply the DRL policy to real world due to the discrepancy between simulation and real world, also known as \emph{reality gap} \cite{realitygap1, realitygap2}. A lot of works were devoted to bridge the reality gap. For example, Domain Randomization (DR) \cite{tobin2017domain, peng2018sim} proposes to train policy in a wide range of environments with various parameters and noises, which makes policy more robust. Additionally, one can also adapt the policy trained in simulator to real-world data in latent space, called domain adaptation \cite{tan2018sim, Junsong2022toward}. Besides, system identification, which tries to identify the physical parameters of real-world explicit or implicit \cite{osi, peng2020learning}, may also help improving the transferring performance. 

By leveraging these advanced skills, DRL has achieved remarkable strides in the fields of 
legged robot locomotion \cite{rma, peng2020learning, lai2023sim},
robotic arm manipulation \cite{AIRL, bettini2023heterogeneous, romero2023Learning},
and wheels robot application \cite{CraftEnv, Zhang2023Abstractions, chen2023mitigating}.
Within the domain of quadruped robots,  
Lee et al. \cite{terrain} and Kumar et al. \cite{rma} developed the teacher-student training paradigm for quadruped robot learning, which has been demonstrated to be highly effective in experiments. 
Lai et al. \cite{lai2023sim} introduced a two-stage training framework called TERT, demonstrating its superior performance across diverse, challenging terrains, showcasing robust task mobility. 
Nahrendra et al. \cite{nahrendra2023dreamwaq} presented a robust framework for quadrupedal locomotion, enabling stable movement across unstructured terrains. 
Peng et al. \cite{peng2020learning} presented a robust framework allowing legged robots to acquire agile locomotion skills by imitating real-world animals’ movements. 
Escontrela et al. \cite{escontrela2022adversarial} presented an effective style reward introducing Adversarial Motion Prior approach into quadrupted robot control.
Similarly, Wu et al. \cite{wu2023learning} formulated a single policy, trained through DRL, to achieve a harmonious balance between robust and agile quadrupedal locomotion. 
Meanwhile, Yu et al. \cite{yu2023multi} developed a learning-based control method to control quadrupedal robots with varying morphologies effectively. 

Notably, the works mentioned above primarily revolve around the premise of the robot's optimal operation, with seldom specific provisions for fault tolerance in the presence of malfunctions. 

\subsection{Fault-Tolerant Quadrupedal Locomotion}
Common robot faults can be categorized into "locked" and "uncontrollable" \cite{chen2022fault}. In the former, the malfunctioning joint becomes immovable and can be used to provide support, while in the latter, the joint becomes freely movable and loses its supporting function. 
This paper mainly focuses on the "uncontrollable" fault scenario, which presents higher challenges and lacks a relevant learning-based framework for quadruped robots.

Numerous studies have delved into alternative strategies, including traditional control theories and other methodologies, to tackle the challenge of fault tolerance in quadruped robots. 
For instance, Cui et al. \cite{cui2022fault} and Chen et al. \cite{chen2022fault} introduced control-based methods to enable quadruped robots to overcome single-joint locking scenarios.
Zhao et al. \cite{zhao2023gait} devised an adaptive fault-tolerant control law to address leg joint actuator faults in quadruped robots.
% They may have presented simulations to validate their approach in a controlled environment.
Allard et al. \cite{allard2023online} and their team have undertaken a series of endeavors rooted in Quality-Diversity that successfully tackled the challenge of damage recovery in real-world hexapod robots. 
This approach, inherent in autonomous learning, showcases significant potential. 

However, a common challenge these methods share is their substantial reliance on task-specific expertise, as they exhibit a heightened dependence on the specification of lower-level behavioral logic. In contrast, the end-to-end learning capabilities of deep reinforcement learning, coupled with the inherent adaptive autonomy, can yield a robust and stable control strategy to address robot faults. 
Okamoto et al. \cite{okamoto2020reinforcement} demonstrated the feasibility of deep reinforcement learning in tackling actuator failure within the ant-v2 environment of the OpenAI Gym \cite{brockman2016openai}. 
Yan et al. \cite{yan2022active} introduced a fault-tolerant reinforcement learning control framework for robotic manipulator joint actuator faults. 
Furthermore, the DRL-based hardware fault-tolerant controller devised by Wu et al. \cite{wu2023learning} for quadruped robot locomotion successfully validates in both simulator and real-world environments. Importantly, their emphasis was on the "locked" fault scenario, as previously highlighted, which is notably distinct from the central focus of our investigation.

% To sum up, research involving reinforcement learning to address "uncontrollable" actuator failure issues and validate them in the real world is currently lacking. Our work aims to fill this void. 

%% file: _preliminaries.tex
\subsection{Reinforcement Learning for Robotic Locomotion}
    \label{sec: PRE}
We conceptualize robotic locomotion as a Markov Decision Process (MDP), delineated by the tuple $(\mathcal{S}, \mathcal A, P, r, \gamma, \mu_0)$, where $\mathcal S$ denotes the robotic state space, containing the robot's proprioception information. 
The symbol $\mathcal A$ signifies the action space, which is described in detail in Section~\ref{sec: architecture}. 
The transition density $P( \cdot | s_{t}, a_t)$ specifies the likelihood of transitioning to various states when taking action $a_t$ in the given state $s_t$. 
We denote the reward function as $r(s_t, a_t)$, while $\gamma$ represents the discount factor and $\mu_0$ represents the initial state distribution. 
Reinforcement Learning (RL) strives to identify the optimal policy $\pi^*: \mathcal S \to \mathcal A$, which maximizes the expected accumulated return, factoring in the discount, throughout decision-making:
\begin{equation}
    \pi^* = \mathop{\arg \max} \limits_{\pi} \mathbb E_{s_0 \sim \mu_0, a_t \sim \pi(\cdot | s_t), s_{t+1} \sim P(\cdot|s_t, a_t)}\left[ \sum_{t=0}^{\infty} \gamma^t r(s_t, a_t)\right].
\end{equation}

\subsection{Adversarial Motion Prior}
The design of reward functions for quadrupedal locomotion is non-trivial and usually needs much domain knowledge and human effort \cite{rma, terrain}.
To address this issue, Adversarial Motion Prior (AMP) \cite{escontrela2022adversarial} is proposed to simplify the reward function design by utilizing some pre-collected expert motions as prior. Specifically, AMP utilizes a similar framework with generative adversarial imitation learning (\textsc{Gail}) \cite{ho2016generative} to discriminate whether a transition $(s_t, s_{t+1})$ is sampled from expert motions. The training objective for the discriminator is defined as:
\begin{equation}
\begin{aligned}
   \mathop{\arg \min}\limits_{\varphi} & \mathbb E_{(s_t, s_{t+1})\sim \mathcal D} \left[(D_\varphi(s_t, s_{t+1} -1)^2 \right] \\
   & + \mathbb{E}_{(s_t,s_{t+1}) \sim \pi} \left[ (D_\varphi (s_t, s_{t+1}) + 1)^2\right]\\
   & \ + \frac{\alpha^{gp}}{2} \mathbb E _{(s_t, s_{t+1}) \sim \mathcal D} \left[ \| \nabla_\varphi D_\varphi (s_t, s_{t+1})\|_2\right].
\end{aligned}
\end{equation}
which contains two least square terms and a gradient penalty term with coefficient $\alpha^{gp}$. The discriminator is then used to calculate the style reward via Equation \ref{eq:style_reward}, which gives a larger reward if the state encountered has a lower probability of being discriminated from the expert dataset by $D_\varphi$.
\begin{equation}
    r_t^{\text{style}}(s_t, s_{t+1}) = \max \left[0, 1- 0.25(D_\varphi(s_t, s_{t+1}) - 1)^2\right].
    \label{eq:style_reward}
\end{equation}
Intuitively, The style function regulates agents' actions with a similar style of reference, which guides the robot into a natural gait. 

\subsection{Decision Transformer}
The Transformer architecture \cite{vaswani2017attention} has garnered significant attention in various domains, including computer vision, natural language processing, and beyond. The attention mechanisms capture the long-range dependencies and relevance, effectively tackling sequential decision-making problems \cite{Muning:176349}. 
Recognizing its potential, Chen et al. abstracted the reinforcement learning (RL) problem into the framework of conditional sequence modeling, subsequently proposing a transformer-based solution known as Decision Transformer (DT) \cite{chen2021decision}. By leveraging the autoregressive model, DT predicts actions based on past state-action sequences and expected returns. 

Consistent with prior research, we employ the GPT architecture \cite{radford2018improving}, comprising several stacked self-attention layers with residual connections and causal masking. 
The input sequences are embedded as tokens containing position encoding.
Subsequently, each token of this sequence is mapped to key, query, and value pairs $k_i$, $q_i$, and $v_i$ through the self-attention layers. 
The resultant output of the self-attention layer is represented as follows:
% \begin{equation}
% \textsc{z}_i = 
% \sum_{j=1}^{n} \text{softmax} \left(
% \frac{\{ \langle q_j,k_{j'} \rangle\}_{j'=1}^n}{\sqrt{\textsc{l}_t}}
% \right)_j \cdot v_j .
% \end{equation}

\begin{equation}
   z_i = \sum_{j=1}^n \mathrm{softmax}(\{\langle q_i, k_{j'} \}_{j'=1}^n)_j \cdot v_j .
\end{equation}

% In the equation above, we use \textsc{l} to express the length of queries and keys to prevent potential symbolic ambiguity. 

In addition, we use position encoding \cite{vaswani2017attention} to model the sequential relevance:
\begin{equation} \ \label{eq: position}
    p_t^i = 
    \begin{cases}
        \cos (\omega_k \cdot t) & \text{if }i=2k\\
        \sin (\omega_k \cdot t) & \text{o.w.}
    \end{cases}, 
    \omega_k = \frac{1}{10000^{2k/d}}.
\end{equation}

%% file: _method.tex
\begin{figure}[htb]
  \centering
  \vspace{10pt}
  \includegraphics[width=0.495\textwidth]{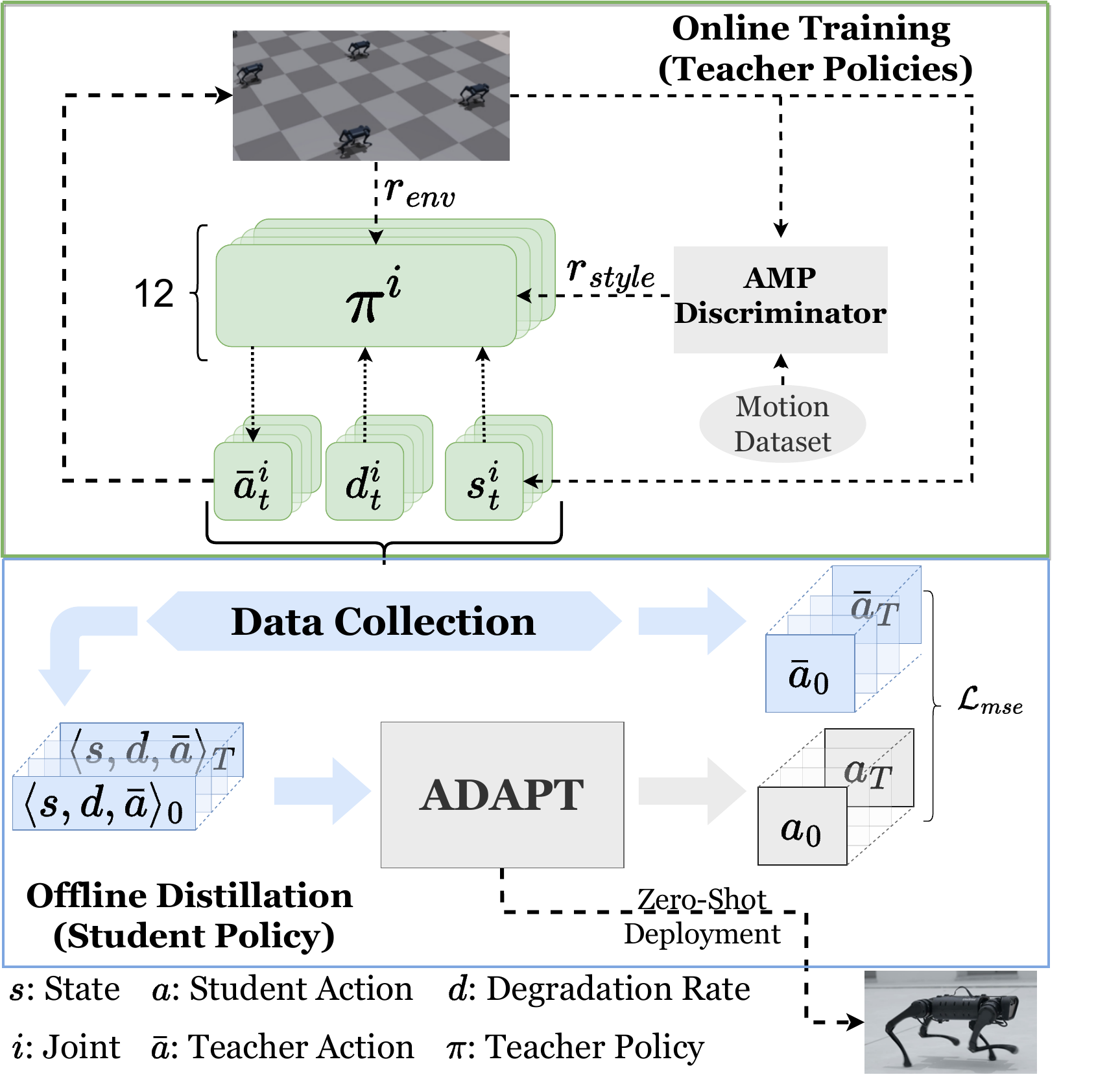}
   \vspace{-15pt}
  \caption{Overall framework of \textsc{Adapt}.
  \textsc{Adapt} starts by training 12 teacher policies separately in simulation. These teacher policies are then used to generate trajectories, which are subsequently utilized to distill a unified transformer-based student policy. Afterward, the student policy is poised for zero-shot deployment to real-world robots. 
}
 \vspace{-10pt}
  \label{fig:framework}
\end{figure}

This section comprehensively explains our novel approach, the Actuator Degradation Adaptation Transformer (\textsc{Adapt}). 
\textsc{Adapt} is devised as a teacher-student framework meticulously designed to incorporate and leverage actuator degradation information. 
The teacher policy within the \textsc{{Adapt}} framework employs actuator degradation information as prior knowledge. 
This utilization of prior information serves the purpose of narrowing down the potential state space during policy exploration. 
On the other hand, the student policy within the \textsc{{Adapt}} framework utilizes actuator degradation information to enhance decision-making processes. 

\subsection{RL-based Control Architecture}
\label{sec: architecture}
The environment is defined precisely as the Markov Decision Process (MDP) outlined in Section \ref{sec: PRE}. 

\minisection{State} 
In the context of this article, our data originates solely from the robot's internal sensory inputs, without incorporating any external sources or additional information such as images or GPS data. 
Consistent with previous designs \cite{pmlr-v164-rudin22a, lai2023sim, yu2023multi}, we use a 48-dimensional vector $s \in \mathcal S$ as the state. This state vector contains both the base and each joint's information (12 joints in total). 

\noindent The base information includes several key variables: 
\begin{itemize}[leftmargin=15pt]
  \item Base linear velocity $v \in \mathbb R^3$. Estimated through comprehensive readings from the IMU accelerometer, the current orientation of the base returned by IMU, and the velocity estimation of the contact leg. 
  \item Base angular velocity $\omega \in \mathbb R^3$. Derived from the IMU gyroscope. 
  \item The body's orientation, represented by the gravity component  $g \in \mathbb R^3$, is calculated based on the base direction measured by the IMU.
  \item Target linear velocity commands $(\hat v_x, \hat v_y) \in \mathbb R^2$ and target angular velocity $\hat \omega_z \in \mathbb R$.
\end{itemize}

\noindent The information for each joint encompasses:
\begin{itemize}[leftmargin=15pt]
  \item Positions $q \in \mathbb R^{12}$ and velocities $\dot q \in \mathbb R^{12}$ can be retrieved through the motor encoders. 
  \item The action $a' \in \mathcal A$ from the previous time step will also be considered.
\end{itemize}

\minisection{Action} 
$a \in \mathcal A$, a 12-dimensional vector, signifies the target position of each joint. 
% Drawing from insights in traditional control theory, position control is deemed safer and more inherently stable than direct torque control. (Dangling Modifier)
Based on insights from traditional control theory, position control is considered safer and more stable than direct torque control.
We subsequently relay the action $a_t$ through a Proportional-Derivative (PD) controller, which generates the desired torque output for each actuator:
\begin{equation}
    \tau^i_{\mathrm{desired}} = K_p (\hat q - q) + K_d (\hat {\dot q} - \dot q) .
\end{equation}
Here, $(K_p, K_d)$ represents the stiffness and damping gains. 
These two hyper-parameters are fundamental components of the PD controller and remain constant throughout our entire training process.

\minisection{Actuator Degradation Rate} 
\label{ADrate}
Inspired by the design concept of EAT \cite{yu2023multi}, we introduce the actuator degradation rate to quantify the degradation extent of quadruped robot joints at the current time step. 
Denoted as $d \in \mathcal D \subset \mathbb R^{12}$, this 12-dimensional vector represents the actuator degradation rate for each joint within the robot's structure. 
% Unlike the corresponding element $e_i$ in EAT, the distinction extends beyond just the task it addresses. 
Notably, $d^i$ is continuous, effectively adapting to diverse fault scenarios during the quadruped robot's operation. 
This continuity enables smooth transfers between different fault states. 
Each component  $d^i \in [0,1]$, where $d^i = 0$ denotes an intact joint, and $d^i = 1$ signifies a joint entirely devoid of force. 
Formally, the actuator degradation can be expressed as:
\begin{equation}
\label{eq:torque}
    \tau^i_{\mathrm{applied}} = \tau^i_{\mathrm{desired}} (1 - d^i).
\end{equation}
Here, $\tau^i_{\text{applied}}$ represents the actual torque applied to joint $i$, while $\tau^i_{\mathrm{desired}}$ is the desired torque on joint $i$. 
%The latter is determined by the PD controller based on the specified action through Equation~\ref{eq:torque}. (Grammarly, For clairty)
The PD controller determines the latter based on the specified action through Equation~\ref{eq:torque}. 
Notably, this torque adjustment directly affects the actual torque output to the joint in both simulated and real-world settings. 
% This manipulation achieves the desired adaptive action as output by the model. 

\minisection{Reward Function} 
Our reward function, denoted as $r: \mathcal{S} \times \mathcal {A} \to \mathbb R$, follows the reward configuration outlined in the article \cite{wu2023learning}. However, in conjunction with experimental knowledge, our approach is further informed by insights from previous work \cite{lai2023sim,pmlr-v164-rudin22a}. 
We have incorporated penalties targeting orientation deviations in the angular velocities of the base's roll and pitch. 
This strategic addition encourages the robot to adopt more stable gaits, particularly when navigating under fault conditions. 
Additionally, we introduce a reward term designed to incentivize the robot to take long footsteps. %tiny torques
Concurrently, we impose penalties for robot collisions. These combined reward and penalty mechanisms contribute to our pursuit of refining the robot's behavior. 
% The reward function used to evaluate the agents is shown in Table \ref{tab:rew}. (Grammarly, Clarity)
Table \ref{tab:rew} shows the reward function used to evaluate the agents. 
\begin{table}[t]
\caption{Reward Function Settings}
\vspace{-10pt}
\label{tab:rew}
\begin{center}
\resizebox{0.9\columnwidth}{!}{
\begin{tabular}{lcc}
\toprule
  \textbf{Term} & \textbf{Reward Formula} & \textbf{Scale} \\ 
 \midrule
        Linear velocity tracking & $\exp\left(-\frac{(\mathbf{v}_{\text{cmd}}-\mathbf{v}_{\text{base}})^2}{0.25}\right) $ & 1.0 \\             
        Angular velocity tracking & $\exp\left(-\frac{(\omega_{\text{cmd}}-\omega_{\text{base}})^2}{0.25}\right) $ & 0.5 \\   
        Unexpected angular velocity& $\omega_{\text{pitch}}^2+\omega_{\text{roll}}^2$ & -0.05 \\   
        Orientation & $x_{\text{gravity}}^2+y_{\text{gravity}}^2$ & -2 \\ 
        Torques & $\|\tau\|^2$ & -0.0001 \\   
        Accelerations & $(\mathbf{v}_{t-1}-\mathbf{v}_{t})^2$ & -2.5e-7 \\    
        Feet air time & $\|\mathbf{t}_{\mathrm{air}} -0.5 \cdot \mathbf{1}\|_1$ & 1.0 \\ 
        Collisions & $\mathbf{1}_{\text{unnatural touchdown}} $ & -0.5 \\ 
        Changes in actions & $\left(a_{t-1}^n-a_{t}^n\right)^2$ & -0.01 \\ 
        Large action & $\|a_t\|^2$ & -0.3\\ 
\bottomrule
\end{tabular}}
% \vspace{-0.05in}
\vspace{-10pt}
\end{center}
\end{table}
% Since our experiment encompassed several parallel environments, the reward output is averaged.

\subsection{Teacher Policy}
Given the high sensitivity of the actuator degradation rate within the task, we train twelve distinct teacher conditional policies. 
Each policy corresponds to one of the twelve scenarios that one joint actuator degrades to some extent. 
 This strategy effectively reduces the policy exploration space, enhancing efficiency. 
Moreover, we adopt the AMP architecture \cite{escontrela2022adversarial} as the foundation of our teacher policy, which regulates the gaits of quadrupedal robots, aligning them with the locomotion patterns observed in expert behaviors. Specifically, in addition to the reward terms outlined in Table \ref{tab:rew}, we introduce a term of style reward \cite{escontrela2022adversarial} (mentioned in Equation~\ref{eq:style_reward}) that evaluates the similarity between the current strategy's gait and that of expert motion.

Let $\bar \pi^i$ denote the $i$-th teacher policy for the scenario where the $i$-th joint degrades. 
The actuator degradation rate at time step $t$ for the $i$-th joint is denoted as $d_{t}^i$. 
The value of $d_{t}^i$ is sampled from the continuous interval [0, 1] at the beginning of each episode. 
Subsequently, the teacher policy $\bar \pi^i$ outputs actions conditioned on the actuator degradation rate $d_{t}^i$ and the corresponding state $s_{t}^i$ :
\begin{equation}
    \bar{a}_{t}^i = \bar \pi^i(s_{t}^i, d_{t}^i).
\end{equation}
The teacher policies are trained via the Proximal Policy Optimization algorithm \cite{schulman2017proximal} and use domain randomization \cite{tobin2017domain} to enhance real-world performance. 

\subsection{Data Collection for \textsc{Adapt} Training}
\label{sec:data}
The meticulous selection of expert data proves pivotal within 
%the context of (Remove by grammarly)
our teacher-student framework for knowledge distillation. 
Our experiment highlights a crucial insight: As the extent of degradation escalates, the robot's gait significantly transforms. 
Therefore, to facilitate the learning of the student, we sample $d_i$ uniformly from $[0, 1]$ at timestep 0 and then adjust it adaptively during data collection: 
\begin{equation}
    \label{eq:joint}
    d_t^i = \begin{cases}
        \mathcal U (d_{t-1}^i, 1) & \text{w.p.~} p, \\
        \mathcal U (0, 0.5) & d_{t-1}^i > 1- \delta, \\ 
        d_{t-1}^i & \text{o.w.}
    \end{cases}
\end{equation}
where $\mathcal U$ denotes the uniform distribution, $\delta$ represents the reset threshold and $p$ signifies the probability of degradation.
%In other words, we increase the degradation rate with a certain probability $p$ at each timestep, and once the degradation rate exceeds the threshold, we reset it to some range.
This mechanism ensures the collected data cover an extensive range of possible degradation scenarios, ensuring a broader skill acquisition by the students. 

Using this mechanism, we collect a set of trajectories encompassing $12 \times N$ trajectories in total, each containing $T$ transitions:
\begin{equation}
\begin{aligned}
    \mathcal T & = \{\mathcal T^i\}_{i=0,1,\dots,11},  \\
    \mathcal T^i &= \{traj_{0}^i, traj_{1}^i, \dots, traj_{N}^i\}, \\
    traj_{k}^i &= \{(s_{k,t}^i, d_{k,t}^i, \bar a_{k,t}^i)\}_{t=0,1,\dots,T}.
\end{aligned}
\end{equation}

Specifically, we set $N = 20000,  T = 500, \delta=0.0001$, $p=0.02$ across our experiments.

\subsection{Actuator Degradation Adaptation Transformer}
Drawing inspiration from Yu et al. \cite{yu2023multi}, 
% who introduced a transformer-based methodology named EAT that accounts for embodiment, 
we propose Actuator Degradation Adaptation Transformer (\textsc{Adapt}). Similar to the embodiment term ${e_i}$ used in EAT \cite{yu2023multi},
\textsc{Adapt} incorporates the actuator degradation rate $d^i$ as an element of the robot's trajectory, serving as a priori information to aid decision-making. 
% In contrast to the earlier approach, where the embodiment term ${e_i}$ was employed in EAT, we replace it with $d^i$, symbolizing the degradation of the quadruped robot's joint actuators. 
This symbolizing enables us to infuse prior information into our experiment, curtailing the exploration space, as discussed in Section \ref{ADrate}. 

% Distinct from prior research employing a global position encoding \cite{chen2021decision, yu2023multi, lai2023sim}, which. 

We chose local position encoding, distinct from prior research employing a global position encoding \cite{chen2021decision, yu2023multi, lai2023sim}. 
To be more specific, in prior methodologies, the position information to be encoded can be denoted as $t$ with its value within 
% the range (Remove by Grammarly)
$[0, T)$,  where $T$ consistently represented the maximum experiment length. 
Such an approach aims to introduce global sequence information to address challenges in other tasks. 
However, we argue that position encoding should be limited to the input sequence's context length. 
Thus, in our \textsc{Adapt}, we set $t \in (0, T_c)$, with $T_c$, the context length, typically much shorter than $T$, 
% specifically configured  (Grammarly)
% configured explicitly to 20, 
aligning more closely with the reality of robot locomotion. 
% This design
% %choice (Remove by Grammarly)
% avoids potential anomalies in time step distribution, especially during prolonged walks, and enhances experimental performance. (Reconstruct by Grammarly)
This design avoids potential timestep distribution anomalies, especially during prolonged walks, and enhances experimental performance.

We adopt an offline training approach similar to \textsc{Gato} \cite{reed2022generalist} to train \textsc{Adapt}, utilizing basic behavior cloning. 
The objective function is formally defined as follows:
\begin{equation}
\begin{aligned}
   & \mathcal{L}(\theta) = \sum_{t=1}^{T} \|\hat a_t , \bar a_t\|_2^2, \\
   (\hat a_1, \hat a_2, \dots, \hat a_T) &\sim \text{Trans}_\theta (s_1, d_1, \bar a_1, s_2, d_2, \bar a_2, \dots, s_T, d_T, \bar a_T).
\end{aligned}
\end{equation}
where $(s_t,d_t,\bar a_t)_{t=1,2,...T}$ is the sequence of transitions sampled in $\mathcal T$. 
\begin{figure*}[htb]
  \centering
  \includegraphics[width=\textwidth]{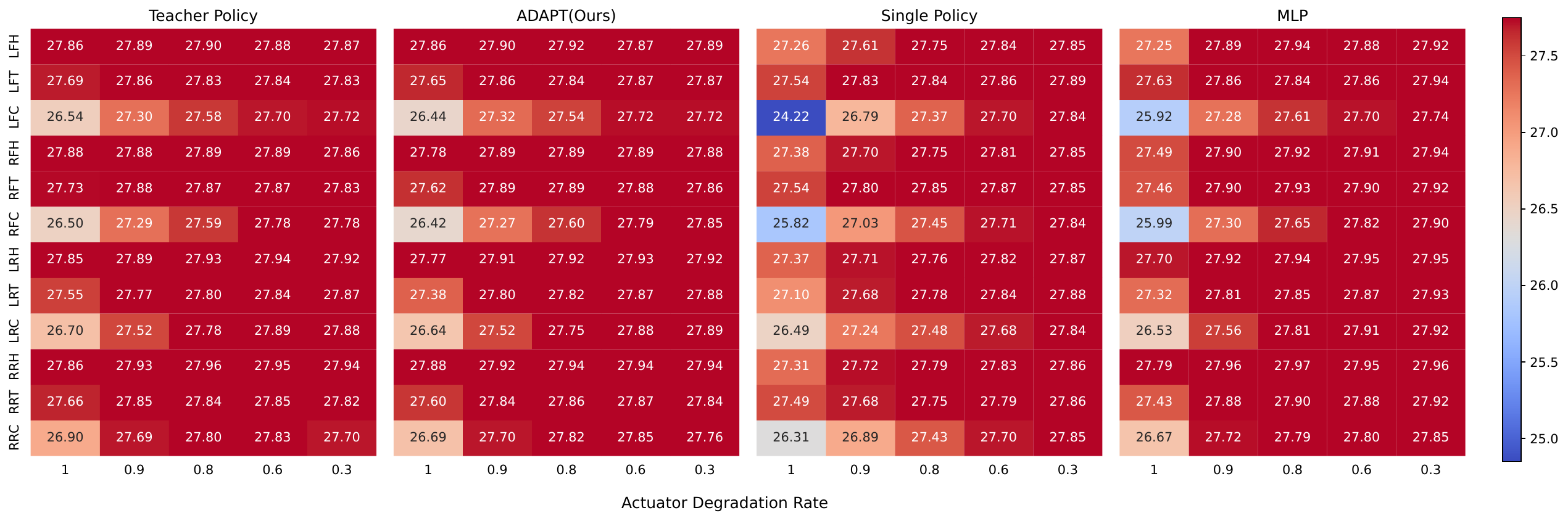}
   \vspace{-10pt}
  \caption{Policy performance over different fault settings (joint, actuator degradation rate). The horizontal axis signifies the actuator degradation rate, with 1 indicating complete damage, while the vertical axis corresponds to the specific joint affected. Each grid in the figure represents the accumulative rewards averaged over 1024 runs in parallel simulation for that specific scenario. Each run randomly sampled the initial state, command speed, and fault occurrence time within specified ranges. The reward recording stopped if the robot fell or 1000 timesteps were collected.}
  \vspace{-10pt}
  \label{fig:heatmap}
\end{figure*}
While GATO uses MLE loss to characterize the distance to experts, we adopt MSE loss instead because of the continuous action space. 
The overall framework of \textsc{Adapt} is shown in Figure \ref{fig:framework} \footnote{The source code has been open-sourced at \url{https://github.com/WentDong/Adapt}.}. 

%% file: _exp.tex
In our cognition, few people are involved in this research topic. Therefore, our experiments focus on answering the following research questions (RQs):
\begin{itemize}[leftmargin=23pt]
    \item[\textbf{RQ1}] How does the model perform under diverse actuator degradation scenarios?  
    \item[\textbf{RQ2}] What are the specific mechanisms by which the model can adapt to changes in actuator degradation?  
    \item[\textbf{RQ3}] What is the performance of our algorithm when deployed to a real-world robot?
\end{itemize}

For ease of presentation, we adhere to the joint naming conventions established by Unitree, which employ a three-letter scheme for each joint. 
The first letter of joint signifies \textbf{F}ront or \textbf{R}ear, the second letter of joint denotes \textbf{L}eft or \textbf{R}ight, and the final letter indicates \textbf{H}ip, \textbf{T}high or \textbf{C}alf. For example, the left front calf is abbreviated as LFC. More experiment settings can be seen in Appendix \ref{appendixA}    

\subsection{Evaluation in Simulator} \label{sec: exp1}
To answer \textbf{RQ1}, we extensively evaluated our model's performance in the simulation environment. The results are shown in Figure \ref{fig:heatmap}.
Due to the lack of existing strategies in this field, we established the following baselines for comparison (to ensure fairness, they were all trained in an identical environment setup): 
\begin{itemize}[leftmargin=15pt]
    \item \textbf{Single Policy.} We tried to learn the same skills with one unified model, denoted as Single Policy, in our experiments. The Single Policy was trained via RL under the conditions corresponding to the sum of all 12 teacher policies until convergence. 
    \item \textbf{MLP.} As an alternative approach, we replaced the transformer architecture in our \textsc{Adapt} with an MLP, using the same training data, resulting in the model denoted as MLP. 
    \item \textbf{Teacher Policy.} We also assessed the performance of the teacher policy under the same conditions, serving as an oracle to evaluate the learning capacity of our student policy. 
\end{itemize}

% We also explored the direct online training of the student policy within the simulation environment, though these outcomes were unable to learn to walk. Hence, we did not present them in the experiments. 
We present a comparative analysis of the performance achieved by four distinct control policies within a simulated environment, visualized through a heat map in Figure \ref{fig:heatmap}. 
The control strategy corresponding to each figure is indicated at the top. 
We did not uniformly select points on the horizontal axis because, in our experiments, when the degradation rate is low, all models show subtle performance variation in response to changes in the degradation rate. 
Therefore, sparsely chosen points are sufficient to represent their performance trends. For more detailed information regarding how performance changes with degradation rate, please refer to Appendix \ref{appendixB}.

The comparison demonstrates \textsc{Adapt}'s adaptability and potential for zero-shot transfer into real-world scenarios marked by diverse joint actuator faults. 
When the actuator degradation rate is low (e.g., below 0.6), all the baselines can achieve satisfactory performance since the models can easily generalize such scenarios through domain randomization. 
However, when the actuator degradation rate rises to some extent, e.g., greater than 0.9, 
both Single Policy and MLP suffer a severe performance deterioration.
Impressively, our model closely matches the Teacher Policy's performance across most scenarios and sometimes even outperforms them. 
Notably, our model's parameters are significantly less than 12 teacher policies, highlighting the pivotal role of knowledge distillation. Specifically, the total parameters of our model are 1.2M, while each teacher policy has 0.4M parameters, adding up 4.8M. 
To make our experiment convincing, we also record the standard deviation of the model's scores under each scenario in the experiment above. 
Our approach has the most stable score, and the detailed experimental results are provided in Appendix \ref{appendixStd}. 

 \begin{figure}[t]
  \centering
  \includegraphics[width=0.48\textwidth]{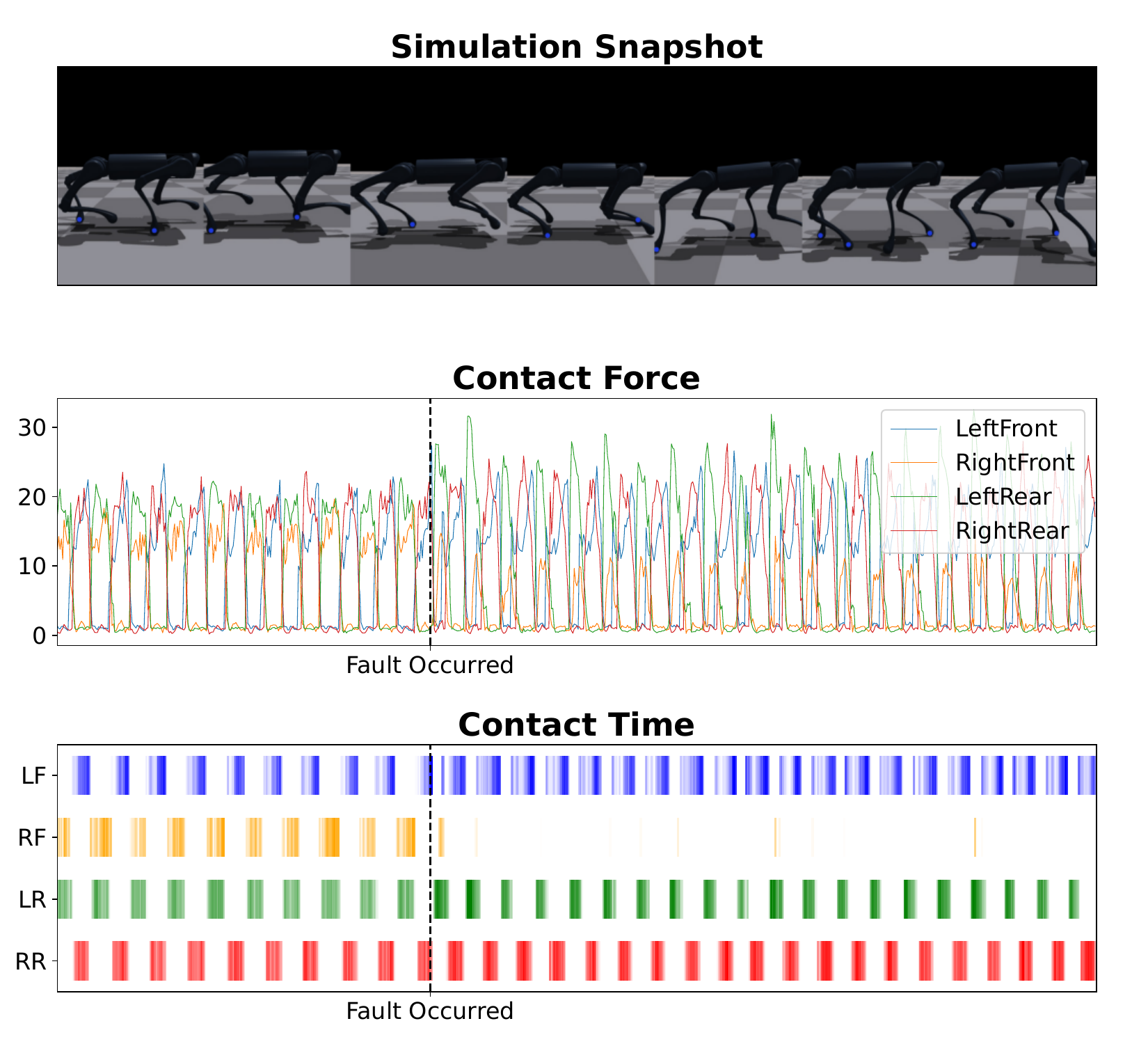}
  % \captionsetup{justification=raggedright,singlelinecheck=false}
  \vspace{-20pt}
  \caption{Illustration of Adaptive gait. Top: The robot's gait changes in the simulator, with ground-contact feet in blue dots. Middle: The L2-Norm of contact force on different feet. Bottom: feet contact time with the ground; darker color represents larger ground contact force.}
  \label{fig:sensor_contact}
  \vspace{-20pt}
\end{figure}
 
Furthermore, the figure indicates that adapting to faults in the calf joints proves more difficult under identical training conditions (rows 3, 6, 9, 12 in Figure~\ref{fig:heatmap}). 
This phenomenon is intuitive since the calf joints must exert larger torque to maintain stability and mobility. 
Therefore, they are significantly more affected by actuator degradation. 
% We believe that the adaptability of these joints to actuator degradation is more crucial. 
We believe that these joints' adaptability to actuator degradation is more crucial. 
Consequently, their performance in severe actuator degradation scenarios in this experiment is worth noting, as it offers valuable insights into the effectiveness of the control policy. 

Meanwhile, exploring the limits of our method's capabilities is also interesting. 
We tested the control performance of our ADAPT when multiple joints failed simultaneously, which means more severe damage or even multiple legs can not work properly at the same time. 
This situation is equivalent to a sudden drastic change in the external environment for the model. 
The experiment results show that our method can adapt to this kind of sudden change to a certain extent. 
Detailed experimental conclusions can be found in Appendix \ref{appendixLimit}.

\subsection{Impaired Adaptive Gait Analysis} \label{sec: exp2}

\begin{figure}[t]
  \centering
  \includegraphics[width=0.47\textwidth]{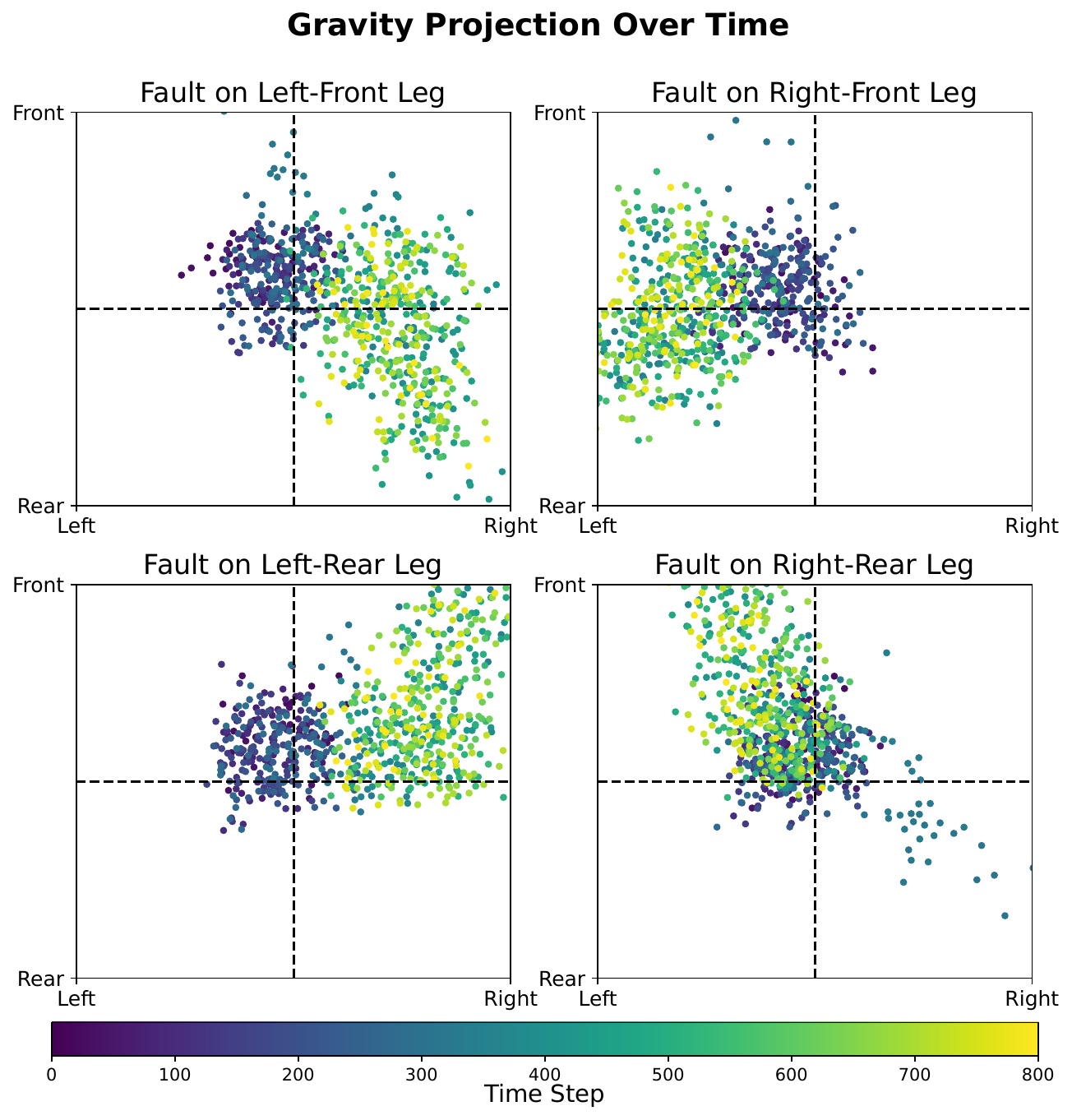}
  % \captionsetup{justification=raggedright,singlelinecheck=false}
  \vspace{-10pt}
  \caption{The projection of gravity components on the robot's plane during robot locomotion, with the robot's forward direction denoted as the positive y-axis and facing right as the positive x-axis. Different colors reflect changes in gravitational force components over time.}
  \vspace{-20pt}
  \label{fig:grv_time}
\end{figure}

We find that our robot adapts to different joint actuator degradation scenarios through what we term "impaired adaptive gait", which describes our robot's capacity to modify its walking pattern in response to varying degrees of joint degradation or impairment. 
To address \textbf{RQ2}, we primarily investigate how our model accommodates regular and impaired adaptive gait patterns in this section. 

Specifically, we command the robot to walk forward at a speed of 1 m/s and then abruptly induce complete degradation in one of its joints. 
We recorded the pressure feedback from its foot sensors and analyzed gait variations based on this force sensing, as depicted in Figure \ref{fig:sensor_contact}. 
To pose a challenge, we set the forward velocity to the maximum linear velocity limit in the training data and select the RFC joint for degradation, as evident from Figure \ref{fig:heatmap}. 
Notably, gait transitions may not be as pronounced in more straightforward tasks.

It can be observed that after the fault occurs, the contact force on the right front (RF) leg quickly decreases, which suggests that the right front leg is dragged forward and cannot provide support. 
In contrast, the left rear leg (LR) and right rear (RR) legs show a significant darkening in color, indicating that they provide more support (which is also evident in the middle image). 
At the same time, the left front (LF) leg has a longer contact time, which means the robot's center of gravity shifts, allowing it to achieve a new stable state under the support of three legs and continue its walking task. 

\begin{figure*}[htb]
    \centering
    \includegraphics[width=\textwidth]{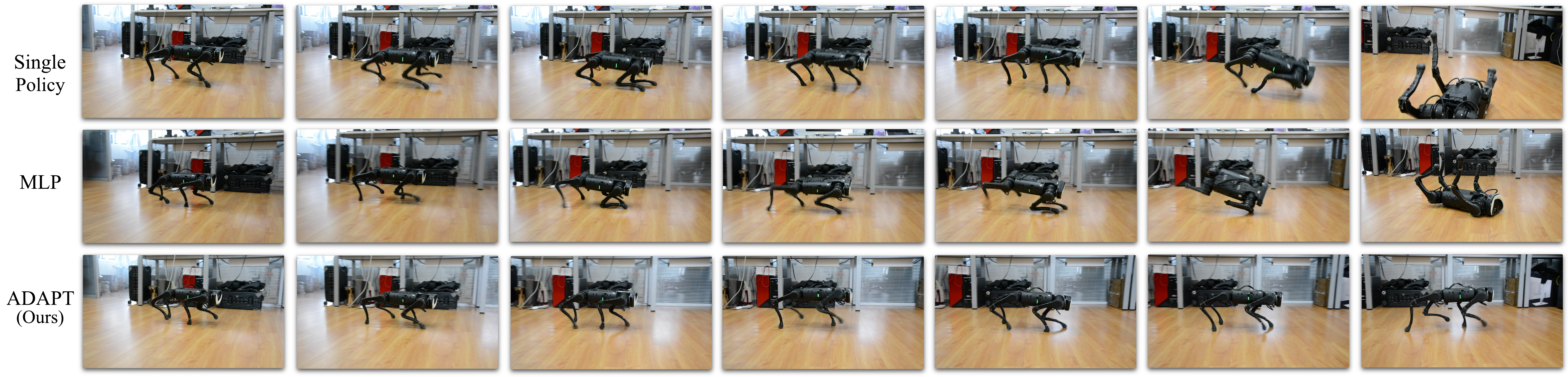}
     \vspace{-20pt}
    \caption{Real-world deployment results of three control models for actuator degradation, with damage restricted to the LFC joint. Images have been mirrored for visual consistency. }
    \vspace{-10pt}
    \label{fig:real-world}
\end{figure*}

Interestingly, we discovered that when gait transitions occur, the robot's center of gravity shifts towards the opposite side of the faulty joint, which is against the intuition that the center of gravity would move towards the direction of the faulty joint due to a lack of support in the direction. 
To verify this, we conducted experiments in a noisy environment, where the command speed was randomly sampled between 0.2 and 1.0, and the faults occurred randomly between 200 and 500 timesteps. We plotted the gravity projection at different timesteps in Figure~\ref{fig:grv_time}. 

It can be observed that at the beginning of each episode, the gravitational force components are initially close to the origin. 
As time progresses and faults occur, their gravity projection gradually shifts toward a particular direction. 
The four plots correspond to severe calf joint actuator degradation scenarios for each leg. 
It can be observed that when a fault occurs, the direction of the center of gravity shift tends to be away from the location of the fault leg. 
In other words, the model adjusts its gait to shift the center of gravity toward a position that facilitates three-legged support. 
This phenomenon is evident across various joint degradation scenarios and becomes more pronounced as the degree of degradation increases. 
For additional experimental results, please refer to Appendix \ref{appendixC}.
This is a strong indicator that our control strategy possesses a distinct impaired adaptive gait%, which is how it adapts to various forms of degradation. 

\subsection{Real World Deployment}

This section primarily delves into addressing \textbf{RQ3}. 
Our work can be successfully transferred to a real-world quadruped robot, demonstrating high adaptability to actuator degradation that aligns with its simulation performance.\footnote{More real-world videos can be found at \url{https://sites.google.com/view/adapt-2023}.} 

To further demonstrate the effectiveness of our proposed model, we directly deploy the models of Single Policy, MLP, and our "\textsc{Adapt}" on a Unitree A1 robot in the real world and induce multiple degradation rates across all joints. 
We simulate actual actuator degradation by attenuating the model's current output torques before sending them to the actuators.
Though successful in low degradation rate scenarios, both Single Policy and MLP fail when the degradation rate is greater than or equal to 0.9.

Figure \ref{fig:real-world} illustrates the performance of three different models when adapting to the left front calf (LFC) joint experiencing a degradation rate of 0.9. \footnote{We did not directly set the degradation rate to 1 because, in such case, the joint actuator would show noticeable damping due to the ongoing motor. However, this does not accurately represent real-world conditions with actuator degradation.} As the figure shows, the Single Policy model exhibits no signs of gait transition. Upon degradation, it attempts to drive the faulty leg similarly to the original gait to maintain balance, leading to imbalance and a fall. Conversely, MLP begins the transition to an adaptive gait only after experiencing a significant deviation in the base to maintain balance. As a result, it still encounters difficulties in maintaining stability. In stark contrast, our "\textsc{Adapt}" method rapidly adjusts the gait by shifting the center of mass to the opposite side (consistent with our analysis in Section \ref{sec: exp2}) and completes the walking task. 

According to real-world experiments, the Single Policy model does not learn the adaptability required for severe actuation degradation, even though it underwent the same training scenarios. 
The MLP baseline lacks the sensitivity to perceive sudden state changes due to the absence of historical information. 
It only transfers to an adaptive gait when there is a significant deviation from its normal state, which is too late to regain balance. 
Our \textsc{Adapt} shows the capability for zero-shot transfer to real-world robot actuator degradation scenarios successfully.

%% file: _conclusion.tex
%The study of the fault-tolerant ability of quadruped robots can play an essential role in quadrupeds’ vigorous future development. 

This paper introduces the \textsc{Adapt} framework to enhance quadruped robots' adaptability to joint actuator faults. 
Our approach utilizes reinforcement learning and a well-designed architecture for smooth transitions between intact and faulty states. 
It can intelligently adapt to various impaired adaptive gaits when an actuator degradation fault occurs. 
Notably, we demonstrate successful zero-shot transferability from simulator to real-world robots, showcasing adaptability across different scenarios. 
In addition, our work has demonstrated the ability to resist sudden environmental changes caused by multiple joint failures to a certain extent. 
As robotics progresses, our framework could further contribute to robust locomotion and broader adaptability across industries. 

Despite this, there is still much room for improvement in our work.
On the one hand, our approach focuses on a specific quadruped robot model, but extending it to various robotic platforms can further demonstrate its generality.
On the other hand, this work focuses more on the motion control part of the quadruped robot, integrating fault detection methods into the control model for a more unified solution is a promising direction.

%% file: _appendix.tex
\section{Training Details} \label{appendixA}
    \subsection{Experiment Settings}
    \minisection{Simulation}
    We implement our model and baselines based on the open-source codebase provided by Rudin et al.  \cite{pmlr-v164-rudin22a}, which ensures thousands of robots simulated in parallel in the environment of IsaacGym simulator \cite{makoviychuk2021isaac}. The URDF file of the A1 robot from Unitree is used during the simulation. 

    \minisection{Hardware} 
    We use the A1 robot from Unitree for all the real-world application experiments, which has 12 actuators (3 motors on each leg) with a maximum torque of $33.5$N.m \cite{unitree}.  

\begin{figure}[htb]
  \centering
  \includegraphics[width=0.49\textwidth]{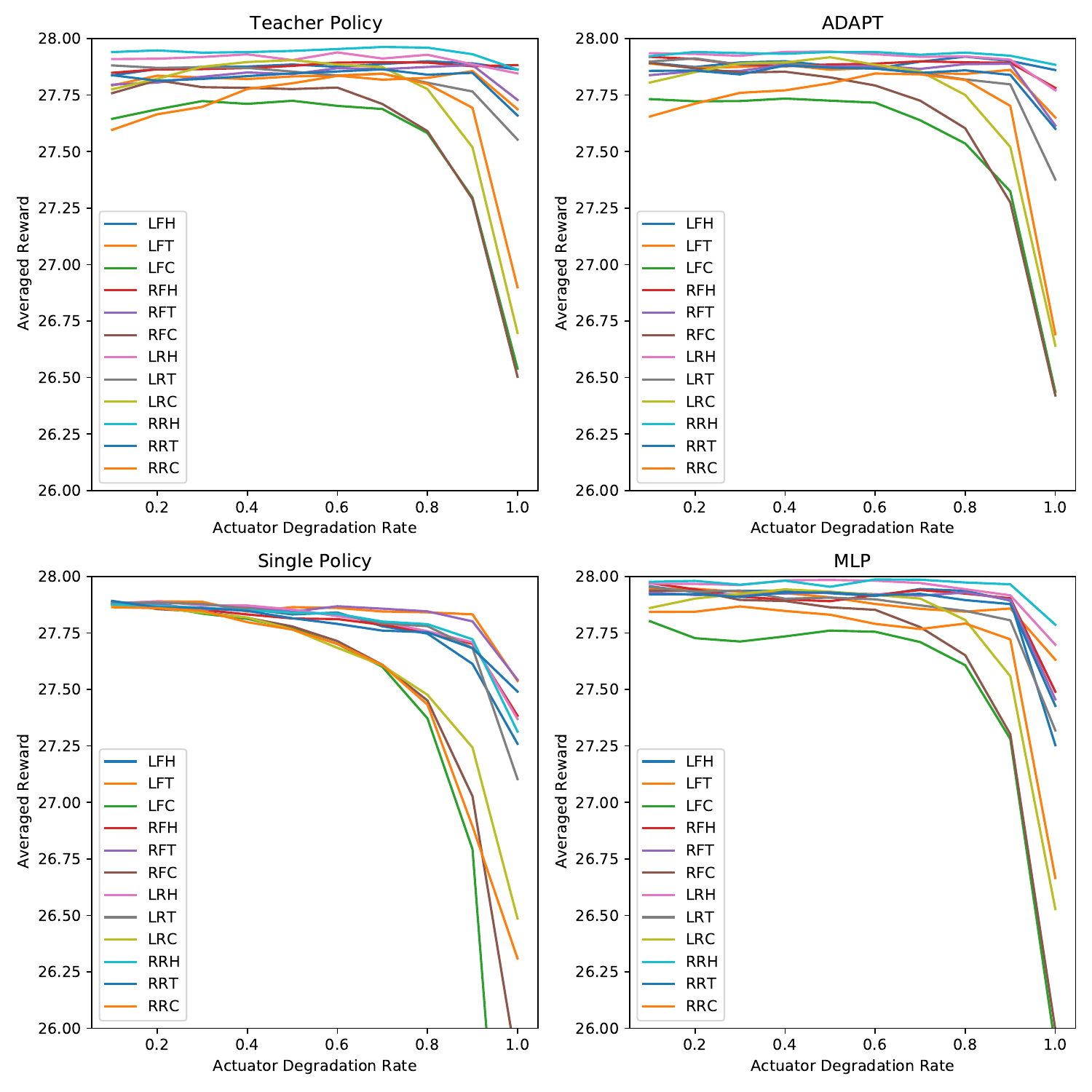}
  % \captionsetup{justification=raggedright,singlelinecheck=false}
  % \vspace{-20pt}
  \caption{The relationship between the performance of the four models and the degradation rate. 
  The horizontal axis represents the actuator degradation rate. The vertical axis shows the model's performance.}
  \label{fig: rew_changes}
\end{figure}
    \subsection{Training Details}
    For the teacher policy, we employ a three-layer MLP (multi-layer perceptron) with $(512,256,128)$ hidden units to encode the states along with the degradation rates and produce the actions. We apply curriculum learning for teacher policy training, which gradually increases the range of the command velocity. 
    Given the physical structure of the A1 robot, it is challenging for the quadruped robot to move laterally or backward with actuator degradation. Hence, we concentrate on forward walking with turning commands sufficient for omnidirectional locomotion \cite{margolis2022rapid}. The maximum walking velocity is set to 1.0 m/s, and the maximum turning angular velocity is 0.5 rad/s.
    
    Moreover, the degradation rate distribution significantly affects the policy's performance. If we sample the degradation rate from a uniform distribution, the training process will be unstable, leading to poor performance under high degradation. Therefore, we use the adaptive sampling mechanism described in Section~\ref{sec:data} for the degradation rate both in the teacher training and data collection phase, which ensures that the degradation changes according to rules that maintain a reasonable distribution. 
    
    We train each teacher policy on one NVIDIA GeForce RTX 3080 for 3000 iterations, which simultaneously takes about 75 minutes, parallelizing 4096 environments. 
    Afterward, we adopt our proposed \textsc{Adapt} framework for student policy training by setting context length $T_c=20$ and the number of blocks $n_b=6$. 
    We train the \textsc{Adapt} model on the collected dataset described in Section \ref{sec:data}, which takes 500 million updates and 20 hours on a single NVIDIA Geforce RTX 3080 GPU. 
    
    \subsection{Domain Randomization}
    Domain Randomization \cite{tobin2017domain} was proposed to address the simulator-reality gap, which randomizes the simulator parameters to expose the model to a wide range of environments during the training process. We adopt Domain Randomization during the training process of teacher policies as well as the collection process of the data set. The details of randomization can be viewed in Table \ref{tab:domain}.
    \begin{table}[htb]
    \centering
    \caption{Domain Randomization Range}
    \begin{tabular}{l c} \toprule
        \textbf{Parameter} & \textbf{Range} \\ \midrule
         Friction & [0.05, 2.75] \\
         Restitution & [0, 1] \\
         Base Mass & [-1, 2] \\
         Link Mass & [0.8, 1.2] \\
         COM Position & [-0.05, 0.05] \\
         $K_p$ & [22.4, 33.6] \\ % 28 * [0.8, 1.2]
         $K_d$ & [0.56, 0.84] \\ % 0.7 * [0.8, 1.2]
         Motor Strength & [0.8, 1.2] \\ \bottomrule
    \end{tabular}
    \label{tab:domain}
    \end{table}

\section{Performance Variation with Degradation Rate} \label{appendixB}
We conducted a series of experiments identical to Section \ref{sec: exp1}, but with a much denser range of degradation rate variations, as shown in Figure \ref{fig: rew_changes}. 

Each subplot illustrates the performance variation of a model under different levels of degradation for 12 joints. The performance is measured by the average reward obtained during experiments.
It's evident that all models exhibit relatively subtle performance changes when the degradation rate is below 0.8. 
However, when it exceeds 0.8, a significant performance decline becomes noticeable. 
This observation further corroborates the discussions in the earlier sections regarding task difficulty. 
Additionally, it's also evident that our \textsc{ADAPT} closely approaches the performance of the Teacher Policy when the degradation rate is low. However, at higher degradation rates, it notably outperforms the baseline. This observation is consistent with the experimental findings presented in Section \ref{sec: exp1}.

\section{Impaired Adaptive Gait Analysis of Various Scenarios } \label{appendixC}
\begin{figure}[htb]
  \centering
  \includegraphics[width=0.48\textwidth]{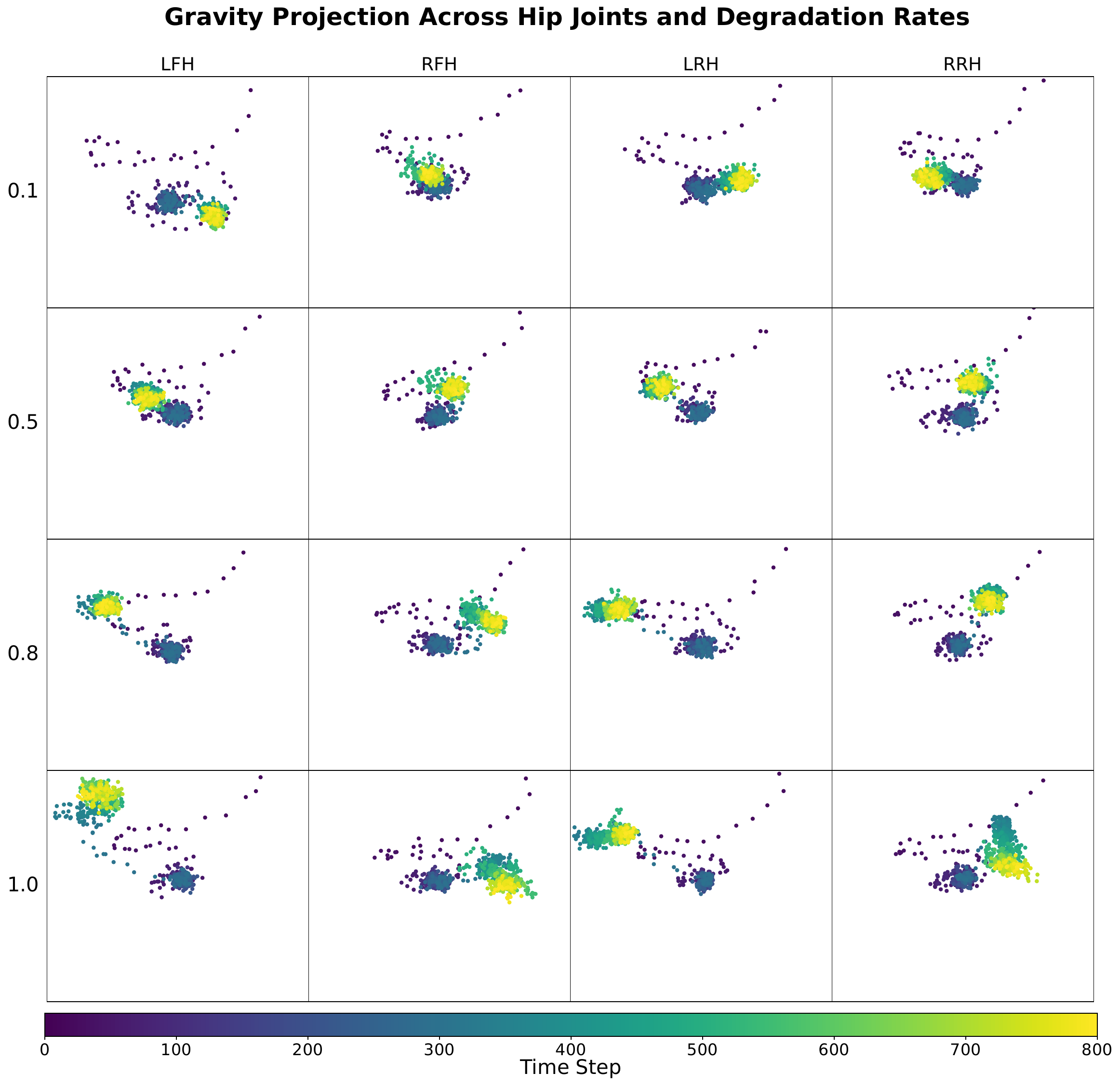}
  \vspace{-20pt}
  \caption{Each column represents a specific damaged joint within the same leg, and each row corresponds to the same degradation rate. }
  \vspace{-10pt}
  \label{fig: grv44_0}
\end{figure}

To comprehensively discuss the gait variations of our control models under different degradation scenarios, we conducted experiments to monitor the shifts in the gravitational projection direction for all joint degradation scenarios. 
The experimental setup and presentation method are consistent with those in Figure \ref{fig:grv_time}. 
To visually emphasize changes in the gravitational projection under different degradation levels, we selected cases with degradation rates of 0.1, 0.5, 0.8, and 1.0 for presentation. 
This choice is based on discussions in Section \ref{sec: exp1} and Appendix \ref{appendixB}. 
\begin{figure}[htb]
  \centering
  \includegraphics[width=0.48\textwidth]{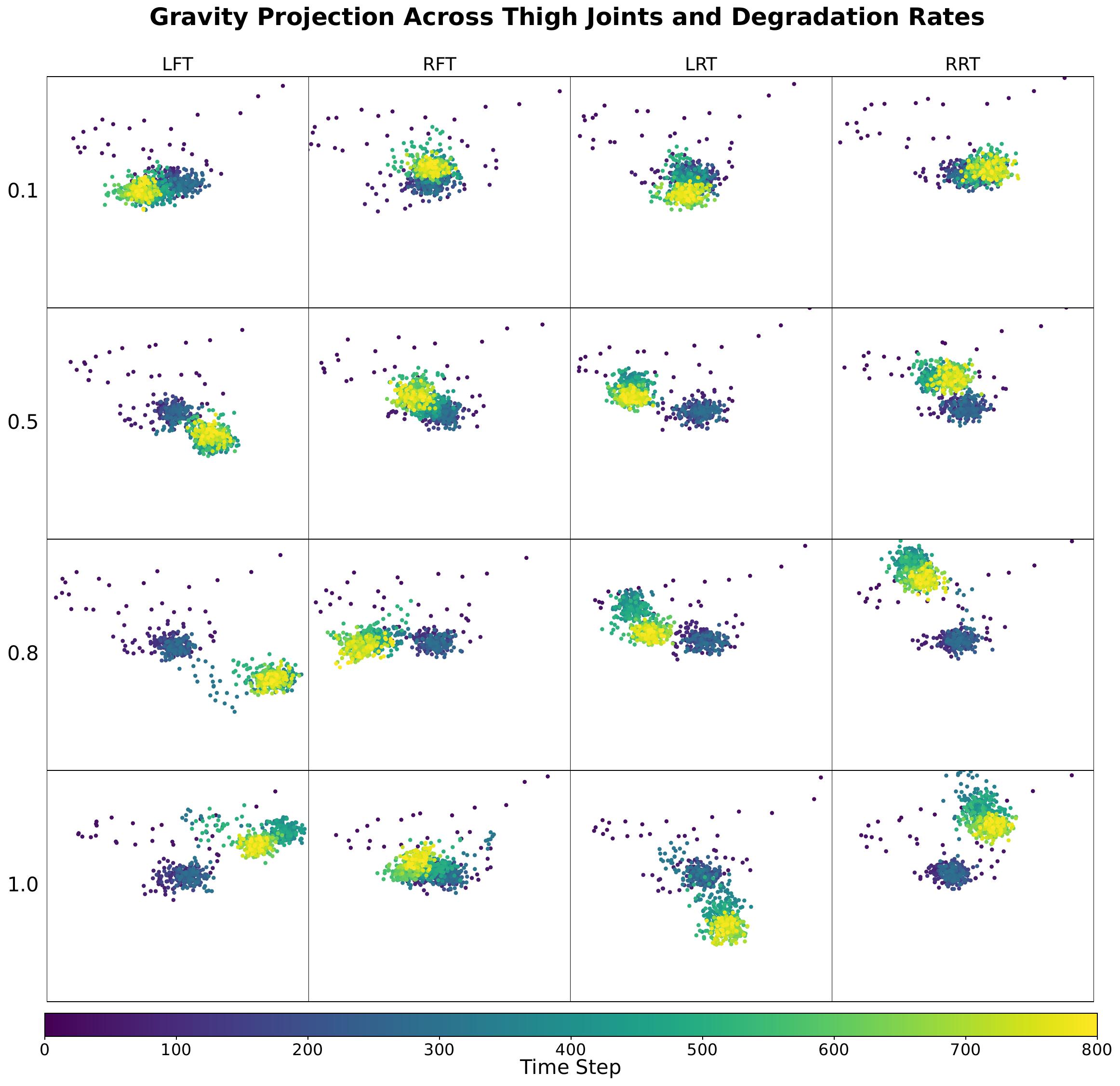}
  \vspace{-20pt}
  \caption{Each column represents a specific damaged joint within the same leg, and each row corresponds to the same degradation rate. }
  \vspace{-20pt}
  \label{fig: grv44_1}
\end{figure}

Figure \ref{fig: grv44_0} illustrates the outcomes of hip joint degradation scenarios. 
From the figures, it is evident that when the hip joint is degraded, as the level of degradation increases, the gravity projection tends to shift more towards the degraded side. 
Seldom reverse shift in the projection of gravity is observed, signifying that the hip joint does not transition to the impaired adaptive gait. 
This is because the hip joint primarily handles leg adduction and abduction and is not as critical for stabilizing the body as other joints. 
Its functionality can naturally be compensated for by other joints. Consequently, it doesn't require a transition into the impaired adaptive gait. 
Such an outcome reflects that the hip joint adaptation is less challenging in the presence of actuator faults, consistent with the conclusions in Figures \ref{fig:heatmap} and \ref{fig: rew_changes}.

\begin{figure}[htb]
  \centering
  \includegraphics[width=0.48\textwidth]{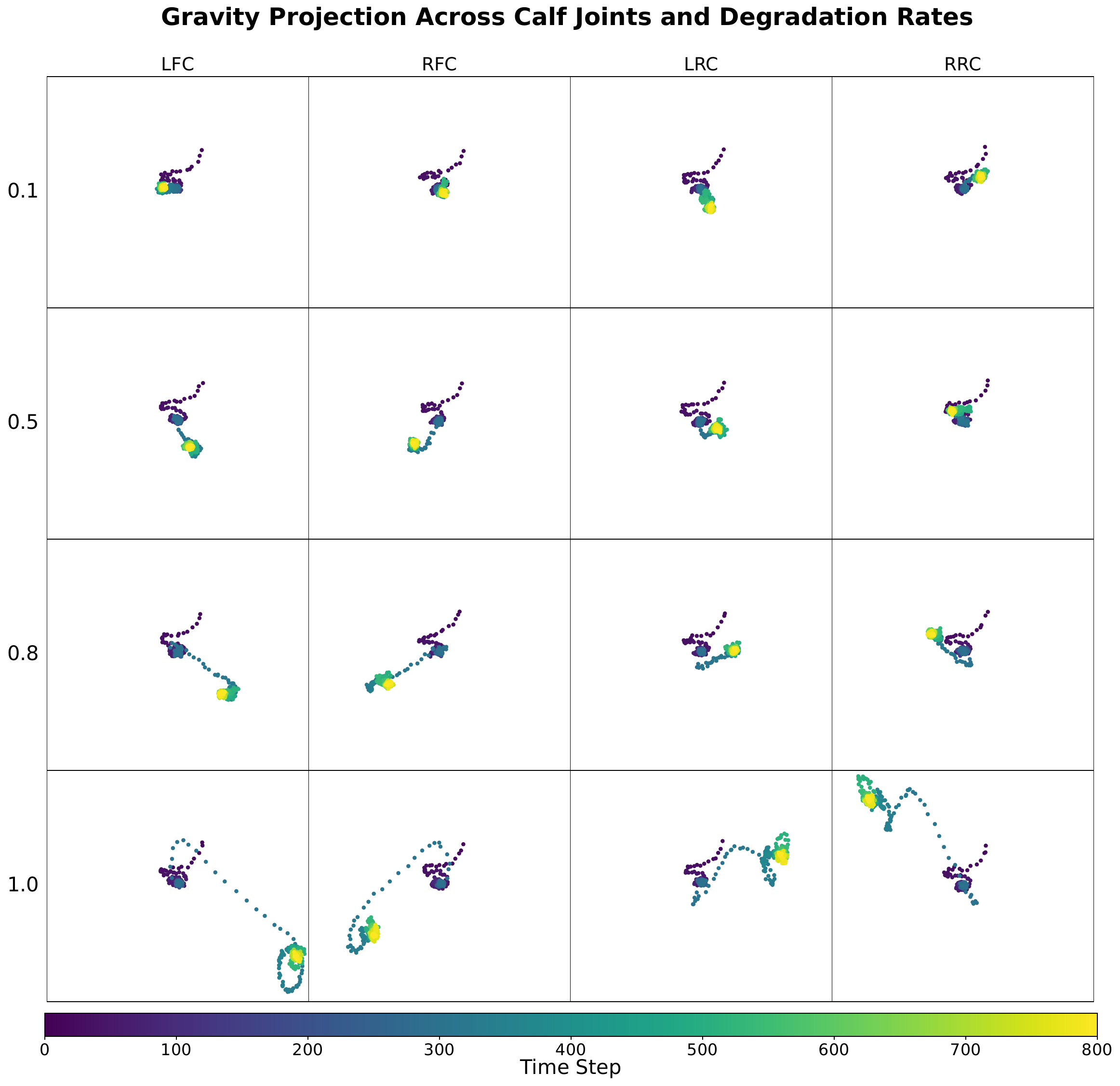}
  \vspace{-20pt}
  \caption{Each column represents a specific damaged joint within the same leg, and each row corresponds to the same degradation rate. }
  \vspace{-20pt}
  \label{fig: grv44_2}
\end{figure}

Figure \ref{fig: grv44_1} and Figure \ref{fig: grv44_2} depict thigh and calf joint degradation scenarios, respectively. 
It's evident that as the degree of degradation increases, both scenarios exhibit a more significant shift in the gravity projection away from the faulty side. 
This suggests that they both show varying degrees of impaired adaptive gait. 
Therefore, the points in Figure 10 appear smaller because, compared to the thigh joint, the shift of gravity projection is more significant when the calf joint is degraded. 
This reflects that the calf joint undergoes more noticeable gait changes. 
This observation aligns with the earlier conclusion drawn from our experiments, emphasizing the more significant challenge of adapting the calf joint to degradation scenarios.

\section{Comparison of standard deviation of scores under different scenarios} \label{appendixStd}
\begin{figure*}[htb]
  \centering
  \includegraphics[width=\textwidth]{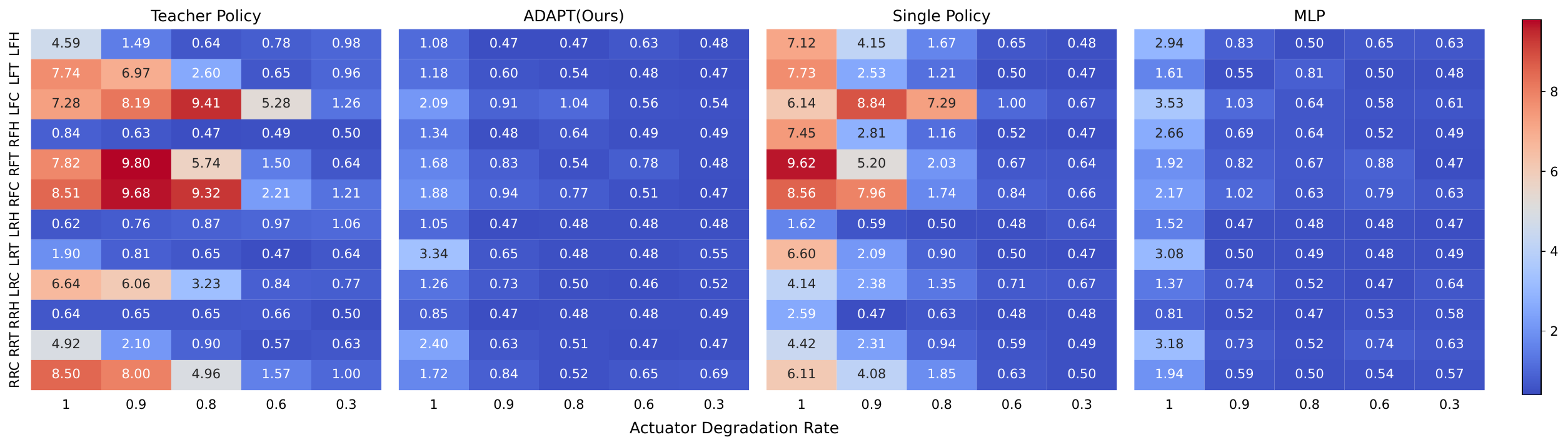}
   \vspace{-20pt}
  \caption{The horizontal axis signifies the actuator degradation rate, while the vertical axis corresponds to the specific joint affected. Each grid in the figure represents the standard deviation of  rewards over 1024 runs in parallel simulation for that specific scenario.}
  \vspace{-10pt}
  \label{fig: heat_std}
\end{figure*}
We experimented with the same settings as in Figure \ref{fig:heatmap} and recorded the standard deviation of scores in all conditions. 
The result is shown in Figure \ref{fig: heat_std}. 
From the figure, it can be seen that as the actuator degradation rate increases, the standard deviation of the model's score becomes larger, indicating that the model's performance is more unstable as it faces more difficult problems. 
However, the Teacher Policy model performs significantly unstable, which we believe is because it has seen relatively few fault situations during training, which limits its generalization ability. 
The Single Policy also exhibits a relatively large standard deviation because it cannot effectively handle fault situations, which is consistent with other experiments' conclusions. 
Both MLP and our \textsc{ADAPT} show relatively small standard deviations, which means that our models perform consistently better than MLP in critical joints.

\section{The scenario of multiple joints getting into fault simultaneously} \label{appendixLimit}
\begin{figure}[htb]
  \centering
  \includegraphics[width=0.48\textwidth]{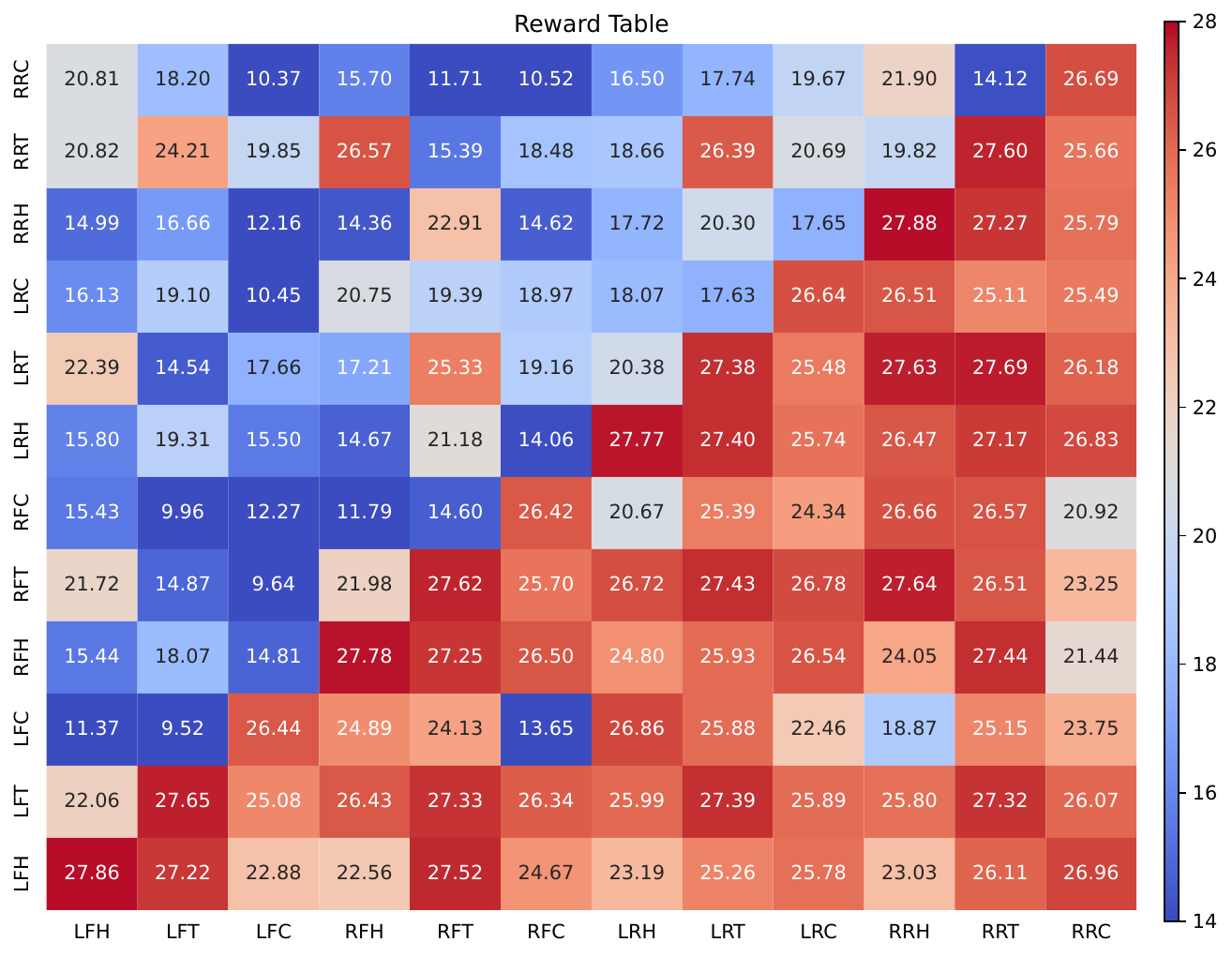}
  \vspace{-20pt}
  \caption{Each grid represents the average reward, with the horizontal and vertical axes representing the specific fault joints. The upper left area represents the case where both joints' actuator degradation rate is 1, while the lower right area represents the case where both joints have an actuator degradation rate of 0.9.}
  \vspace{-10pt}
  \label{fig: limit_2}
\end{figure}
Due to the completely different state space changes caused by multi-joint faults compared to single-joint faults, it can be considered as a completely new task. It should be noted that the above fault scenarios are rarely encountered in real-world deployments. These experiments aim to demonstrate that our method can directly generalize to adapt to certain multi-joint fault scenarios without fine-tuning. 

\begin{figure}[t]
  \centering
  \includegraphics[width=0.48\textwidth]{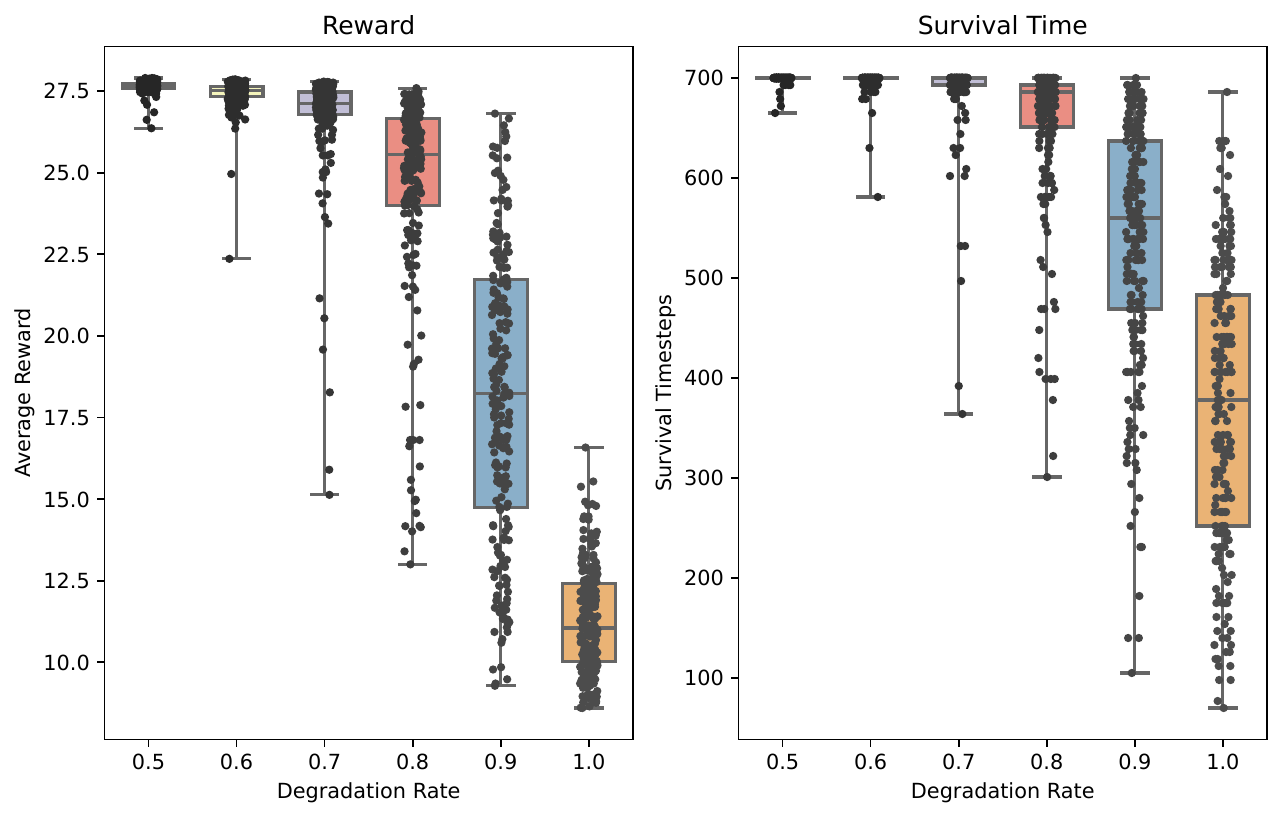}
  % \vspace{-10pt}
  \caption{The left graph represents the average reward obtained from 512 parallel environments, and the right graph represents the time taken by the robot to maintain its stability after encountering actuator degradation. }
  % \vspace{-10pt}
  \label{fig: limit_3}
\end{figure}

\subsection{Dual-Joint Actuator Degradation}
Figure \ref{fig: limit_2} represents the performance of a robot when encountering simultaneous failure of two joints measured by the average reward obtained from 1024 parallel experiments in the simulator. This figure depicts the robot's performance when encountering simultaneous faults in two joints. As can be seen from the figure, even when 90\% of any two joints are damaged, our model can still achieve good average rewards in more than 90\% of the cases. 

\subsection{Three-Joint Actuator Degradation}
We recorded the average reward and the time taken for the robot to maintain its balance after any three joints occurred actuator degradation simultaneously in 512 experiments. We plotted the distribution in Figure \ref{fig: limit_3}. It can be seen that when the actuator degradation rate reaches 70\%, the average reward performance of the model remains at a high level. When the actuator degradation rate reaches 80\%, the average reward decreases, but the survival time does not change significantly, with only an average 6.5\% decrease compared to the original performance. This means that in this situation, the model can still maintain its stability but may not be able to strictly follow commands.

\begin{figure}[b]
  \centering
  \includegraphics[width=0.48\textwidth]{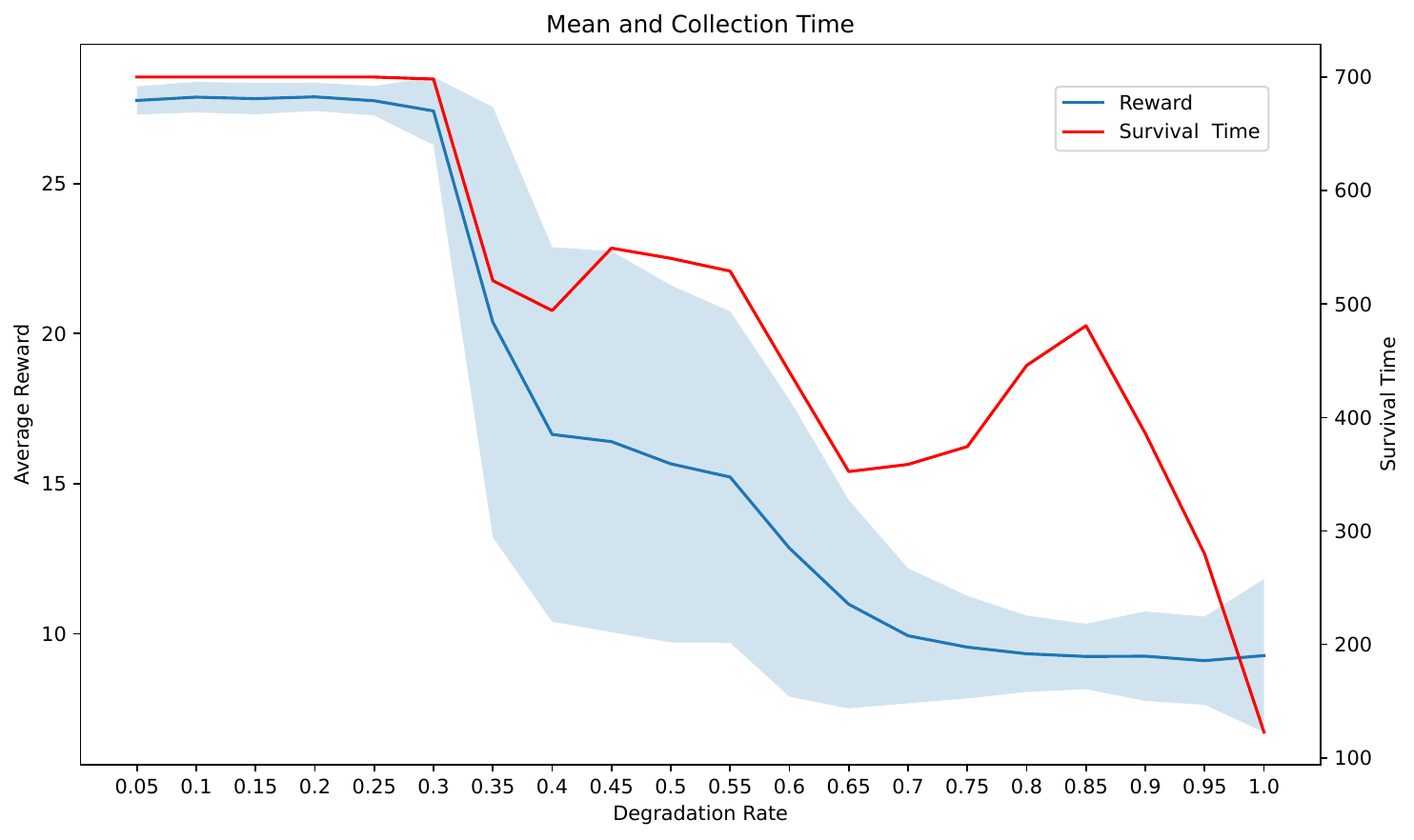}
  % \vspace{-10pt}
  \caption{The horizontal axis represents the actuator degradation rate, with the blue line representing the average reward and the blue shadow reflecting their variance. The red line represents the time taken for the robot to maintain its balance after the failure occurs.}
  % \vspace{-20pt}
  \label{fig: limit_all}
\end{figure}

\subsection{All Joint Actuator Degradation}
Finally, we tested our model's reward and the time taken to maintain balance after all joints simultaneously experienced a certain degree of malfunction. The data was obtained by averaging results from 1024 parallel experiments and shown in Figure \ref{fig: limit_all}. It can be seen that when the degradation level of all actuators is below 25\%, the model performance is not significantly affected, with a decrease of less than 2.5\% in performance compared to the original level. However, as the actuator degradation rate increases, the robot's survival time fluctuates, which is because when the actuator degradation rate increases to a certain level, the robot gives up its movement ability and prioritizes maintaining its standing state. 

Let us consider actuator health as an external environmental indicator. Our model performance will not be significantly affected as long as the change in this indicator is less than or equal to 25\%. However, further exploration is needed to determine the specific boundaries of the model's capabilities.